# Multimodal Magic: Elevating Depression Detection with a Fusion of Text and Audio Intelligence

Lin(Lindy) Gan, Yifan Huang, Xiaoyang Gao, Jiaming Tan, Fujun Zhao*, Tao Yang*
Department of Information & Intelligence Engineering, University of Sanya, Sanya, 572022, China
Academician Workstation of Chunming Rong, University of Sanya

ABSTRACT

This study proposes an innovative multimodal fusion model based on a teacher-student architecture to enhance the accuracy of depression classification. Our designed model addresses the limitations of traditional methods in feature fusion and modality weight allocation by introducing multi-head attention mechanisms and weighted multimodal transfer learning. Leveraging the DAIC-WOZ dataset, the student fusion model, guided by textual and auditory teacher models, achieves significant improvements in classification accuracy. Ablation experiments demonstrate that the proposed model attains an F1 score of 99.1% on the test set, significantly outperforming unimodal and conventional approaches. Our method effectively captures the complementarity between textual and audio features while dynamically adjusting the contributions of the teacher models to enhance generalization capabilities. The experimental results highlight the robustness and adaptability of the proposed framework in handling complex multimodal data. This research provides a novel technical framework for multimodal large model learning in depression analysis, offering new insights into addressing the limitations of existing methods in modality fusion and feature extraction.

## I  INTRODUCTION

Depression is a significant global health concern that affects millions of individuals across various demographics, leading to considerable social, economic, and health-related impacts. According to the World Health Organization (WHO), depression is one of the leading causes of disability worldwide, with over 264 million people affected.[1] The condition is associated with decreased productivity, increased morbidity and mortality, and immense personal and societal burdens. Despite its prevalence, depression often remains underdiagnosed and undertreated due to various factors, including stigma and a lack of accessible and effective treatment options. Moreover, the heterogeneity of depression, manifesting through a diverse range of symptoms and severity levels, complicates its early detection and management.

In recent years, the intersection of clinical psychology and technology, particularly in the realms of machine learning (ML) and deep learning (DL), has opened new vistas for understanding and diagnosing depression. These technologies have demonstrated potential in identifying intricate patterns within large datasets, including electronic health records (EHRs), social media interactions, and wearable device data, offering insights into individual and collective mental health trends.[2] However, traditional machine learning approaches for depression detection primarily focus on single-modal data analysis, such as extracting linguistic features from text or acoustic features from audio, which often fail to capture the complex interplay between different

modalities.[3][4][5] For instance, while linguistic features like sentiment polarity and topic models can provide valuable insights into depressive states, they may overlook critical cues embedded in speech patterns or physiological signals.[6][7] Similarly, traditional classifiers like Support Vector Machines (SVM) and Random Forests (RF) lack the capacity to model high-dimensional, multimodal relationships effectively.[8][9]

The advent of Large Language Models (LLMs), has introduced groundbreaking opportunities in mental health research. These models, trained on extensive corpora of text, can generate human-like text, comprehend context, and engage in meaningful dialogues. In the context of depression, LLMs have shown promise in tasks such as initial screening through linguistic marker analysis and providing scalable, personalized support.[10][11] However, despite their potential, LLMs are primarily limited to text-based data and struggle to integrate other modalities, such as audio or physiological signals, which are crucial for a comprehensive understanding of depression.[12][13] Furthermore, challenges related to data privacy, model interpretability, and the need for large, annotated datasets continue to hinder their widespread adoption in clinical settings.[14]

To address these limitations, we propose an innovative multimodal fusion model based on a teacher-student architecture, designed to enhance depression classification by effectively integrating textual and auditory data. Unlike traditional single-modal approaches, our model leverages the complementary strengths of text and audio modalities through multi-head attention mechanisms and weighted multimodal transfer learning. By utilizing the DAIC-WOZ dataset, the student model is guided by separate teacher models for text and audio, enabling it to learn robust representations of depression-related features from both modalities. This approach not only overcomes the limitations of single-modal methods but also addresses the challenges of modality weight allocation and feature fusion in existing multimodal frameworks.[15][16]

Experimental results demonstrate the efficacy of our approach, with the proposed model achieving an F1 score of 99.1% on the test set, significantly outperforming traditional unimodal and multimodal methods. Ablation studies further validate the contributions of the teacher-student architecture and weighted hybrid loss function in enhancing model performance. Our findings highlight the potential of multimodal large models in depression analysis, offering a novel technical framework that addresses the limitations of existing methods in feature extraction, modality fusion, and generalization capabilities.[17][18]

The contributions put forward in our paper can be summed up as follows:

a. We propose a Teacher-Student Architecture-based Model Fusion Network, which is trained under the guidance of two teacher models. This model comprehensively considers the distinctive contributions of text and audio in the classification and recognition of depression, thereby enhancing the accuracy of the final fused model predictions.

b. We pioneer in introducing a model fusion network with the teacher-student architecture to boost the performance in depression classification tasks.

c. In training the student model, we integrated feature fusion (text and audio) and employed attention mechanisms to calculate the feature weights of different modalities, attaining optimal fusion results. (Refer to Equations 1-5)

d. The $L_{total}$ in the model training consists of KL divergence and cross-entropy, with weights used to balance the contributions of these components during the training process. (Refer to Equation 6)

## Ⅱ  RELATED WORK

In our research, we utilized Python 3.9 and PyTorch 2.5.0 for implementing our models. The computational experiments were conducted on NVIDIA 4070 GPUs with 32GB of memory, alongside Intel i7-14700K CPUs. This high-performance computational setup enabled efficient training and testing of both text and audio models, as well as subsequent model fusion tasks.

*DATASETS*

The DAIC-WOZ database[20][21], a segment of the Distress Analysis Interview Corpus (DAIC), primarily encompasses clinical interview recordings aimed at assisting in diagnosing mental health issues such as anxiety, depression, and Post-Traumatic Stress Disorder (PTSD). This database contains 189 data samples, numbered from 300 to 492. Such interview data are utilized to train computational agents capable of autonomously conducting interviews and identifying mental illnesses through verbal and non-verbal indicators. This dataset includes audio and video records as well as extensive questionnaire responses. A notable feature within the corpus is the inclusion of interviews conducted by 'Ellie', an animated virtual interviewer controlled by a real interviewer from another room. All data have been transcribed and meticulously annotated for both verbal and non-verbal features.

In our experiments, audio data from the DAIC-WOZ database were transformed into text format. These text data were subsequently subjected to thorough cleaning and feature extraction to ensure data quality and usability. Utilizing these preprocessed data, we implemented downstream tasks for information extraction and sentiment analysis. This process not only enhanced the utility of the data but also provided significant technological support for the automated identification and assessment of mental health conditions. Through these advanced analyses, we are able to more deeply understand patients' psychological states, thereby advancing the diagnosis and treatment processes for mental health disorders.

## Ⅲ  METHODOLOGY

We chose BERT-base-uncased as the initial student model. On this basis, we processed audio and text features via a fully connected layer to build a multimodal student model. The model design is presented in Figure 1. During the initial unimodal training, we utilized the DAIC-WOZ dataset, and the detailed methods and steps are elaborated in sections 3.1-3.3.

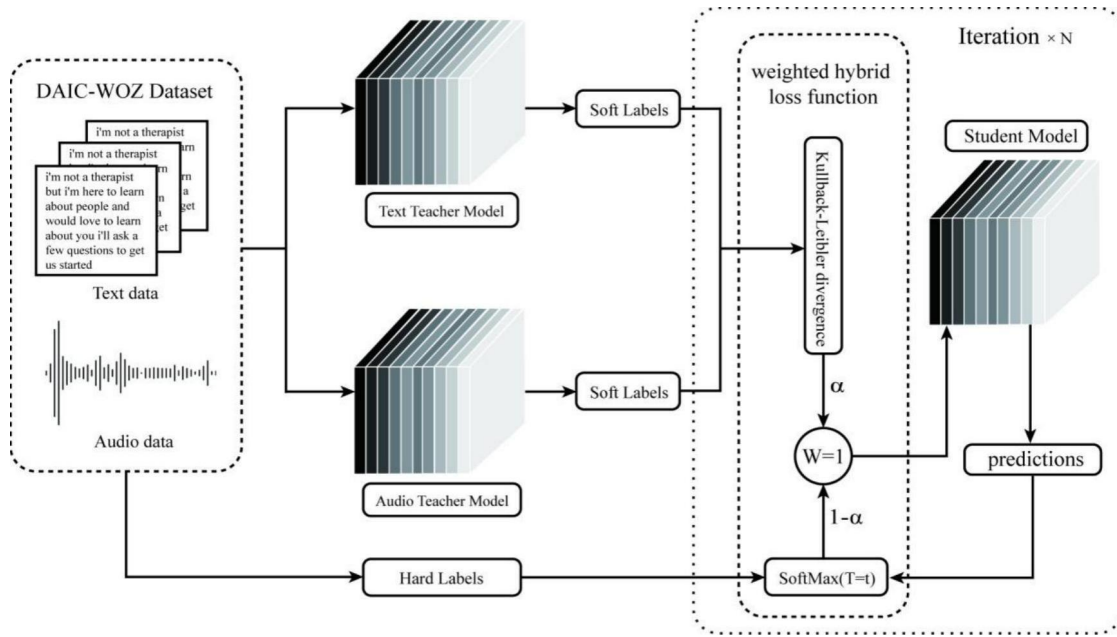

Figure1.Our Model framework.

3.1 Text Modal

In this study, we designed a text classification approach to predict depression based on textual data, utilizing the large language model, Llama, customized with LoRA for enhanced fine-tuning capabilities.

*Data Preparation*

After initially cleaning the text, we removed all questions posed by the virtual character Ellie, retaining only the participants' responses. These responses were then consolidated. The merged text reflects the patients' direct reactions, thereby enhancing our understanding of the participants' perspectives. The consolidated text was processed for tokenization using the tokenizer from the AutoTokenizer class, which is pre-configured for the Llama model. Each segment of text was encoded with special tokens, padded to a maximum length of 512 tokens, and truncated as necessary to ensure a consistent input size for the model.

*Text-model Training*

The core model, the Llama Model, is adapted by appending a linear classification layer, which utilizes the [CLS] token's embedding from the last hidden state of the base model. This token embedding aims to encapsulate the contextual representation of the entire input text, which is then used to determine the probability of depression. Fine-tuning involved the integration of the LoRA technique, which selectively modifies key components of the model's transformer layers — specifically, the query (q_proj) and value (v_proj) projection matrices. LoRA utilizes low-rank matrices with a specified rank of 8, and an alpha scaling factor of 32, offering a balance between adaptability and computational efficiency. This approach allows for significant modifications to the model's behavior with minimal adjustments to its parameters, enhancing its sensitivity to nuanced patterns indicative of depression without the extensive re-training typically required for full model fine-tuning.

Finally, we employ the AdamW optimizer during the training process to maintain a low learning rate, which helps prevent overfitting while adapting the pre-trained model to a specific task. Cross-entropy loss is used to measure the difference between the predicted and true labels, with the model being optimized over 20 epochs. Evaluation is conducted on a separate test set to ensure the robustness and generalization of the model.

### 3.2 Audio Modal

Our audio model architecture is based on a Bidirectional Long Short-Term Memory (BiLSTM) model for training, classifying participants in the DAIC-WOZ dataset.

*Data Preparation*

This study randomly extracts 150 audio clips with emotional labels from the DAIC-WOZ dataset, each resampled to a standard rate of 16,000 Hz to ensure consistency in data processing. During the experiment, finite impulse response (FIR) filters were set to 7,000 Hz to reduce high-frequency noise and enhance the clarity of speech signals. Voice Activity Detection (VAD) algorithms were employed to isolate voice clips and extract Mel Frequency Cepstral Coefficients (MFCC) features. To ensure consistency of input data, each audio clip's sequence length was adjusted to a fixed value (e.g., 60 frames), and they were batched in fixed sizes.

*Audio-model Training*

This study employs a Bidirectional Long Short-Term Memory network (BiLSTM) as the main structure for training. BiLSTM consists of forward and backward LSTM units that process audio features in sequence order and reverse order, respectively. It concatenates the hidden states at each time step from both directions to obtain a final, comprehensive representation. This bidirectional approach better captures the contextual information within the audio sequences. The output from the BiLSTM is mapped to output categories through a fully connected layer, completing the task of depression classification.

After model construction, Mini-batch Stochastic Gradient Descent (SGD) is used for parameter optimization, dividing the training data into several batches for forward and backward propagation. Mini-batch training effectively reduces memory usage while enhancing the model's generalization capabilities through random gradient descent. At the end of each epoch, validation metrics are evaluated, and the learning rate or model parameters are adjusted as needed. The study compares the Adam and AdamW optimizers using the cross-entropy loss function, where AdamW introduces weight decay to better manage overfitting issues.

To further optimize classification performance, this study applies a dynamic learning rate adjustment strategy, using the ReduceLROnPlateau method. This method automatically adjusts the learning rate when the validation loss no longer declines, thus achieving finer optimization during the training process.

*Audio-model Quantization*

Quantization technology is a method that reduces the storage and computational demands of a model by compressing its parameters from high to low precision. In this study, the BiLSTM model was processed using 8-bit quantization, initializing quantization parameters such as scaling factors and zero points, defining the quantization range, and converting floating-point weights into integer values through normalization and clipping. Quantization-aware training (QAT) method was adopted, which simulates the quantization process and computes gradients, enabling the model parameters to adapt to the effects of quantization, thereby minimizing precision loss. Ultimately, this study produced a quantized weight matrix that significantly reduced computational and storage costs while maintaining the performance of the model.

3.3 Teacher-Student Architecture-based Model Fusion Network

The Teacher-Student architecture is an effective multimodal transfer learning approach that achieves model capability transfer through pre-trained teacher models as well as student models trained on a specific dataset. The student model achieves cross-modal transfer of model capabilities by mimicking the output of the teacher model in both modalities, optimizing learning efficiency and reducing resource requirements. This approach allows for the efficient integration of models from different modalities while reducing the need for large amounts of labeled data.

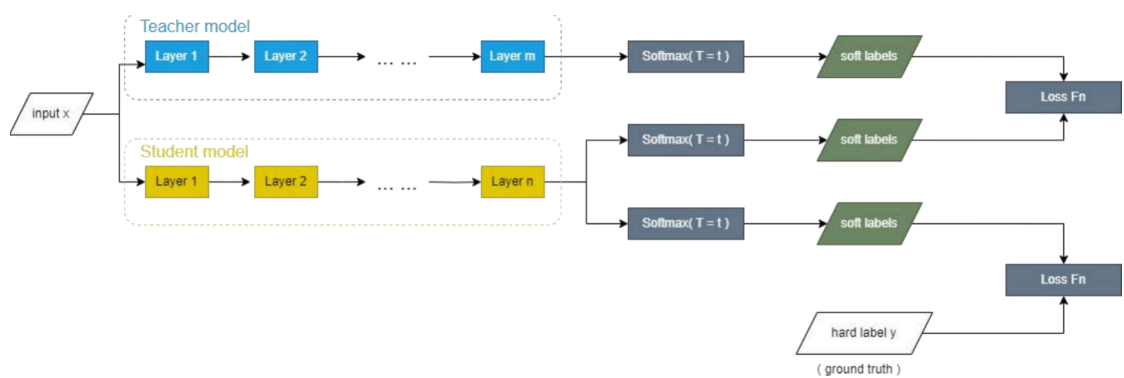

Figure2 .Generic Teacher-Student Framework.

*Teacher Model*

In our research, the pretrained TextTeacher and AudioTeacher serve as the teacher models for the text and audio domains, respectively. These teacher models provide predictions for their respective modalities. Their outputs are used as targets in the model fusion process, guiding the student models to learn richer and more robust feature representations.

*Student Model*

In the experimental section, we developed a StudentModel that combines the experiences of text and audio models by learning the label probabilities from two TeacherModels. In this study, the base BERT model (bert-base-uncased) is selected as the initial StudentModel. In the StudentModel, text feature extraction relies on the BERT model, while audio feature extraction is performed by the Wav2Vec2 model. By merging the output features of BERT and Wav2Vec2, and processing them through a fully connected layer, the StudentModel integrates information from both text and audio modalities to enhance the processing capabilities of multimodal information.

*Attention Mechanism Feature Fusion*

In recent years, a large number of multimodal feature fusion methods have emerged, which can mainly be classified into two types: early fusion and late fusion. Early fusion techniques combine the features of different modalities through concatenation or weighted summation before feeding them into a classifier, late fusion techniques, conversely, model different modalities separately and then fuse the predictive results from each modality. However, these methods frequently overlook the interactions between modalities and have difficulty capturing the associations among them.

To better capture the associations between modalities, we propose a fusion method based on the attention mechanism. Specifically, we map textual and audio features into the same latent space, concatenate them, and then compute attention scores through a linear layer. These scores are transformed into attention weights via a softmax function, and the concatenated features are consequently weighted and summed to generate the final fused feature.

Specifically, let the text feature be $x_t \in R^{d_t}$, and the audio feature be $x_a \in R^a$, we first map them to the same latent space:

$$h_t = w_t x_t + b_t \quad (1)$$

$$h_a = w_a x_a + b_a \quad (2)$$

$d_h$ represents the dimensionality of the latent space, while $d_t$ and $d_a$ correspond to the feature dimensions of visual and textual modalities, respectively. Subsequently, the latent space representations of textual features $h_t$ and audio features $h_a$ are concatenated.

$$h = [h_t, h_a] \in R^{2 \times d_h} \quad (3)$$

Then, we obtain attention scores through a linear layer:

$$e = W_e \cdot h^T + b_e \quad (4)$$

The function of $W_e \in R^{1 \times d_h}$ is to learn the weights representing the correlations among modal features, whereas $b_e \in R$ acts as the bias term. Following this, attention weights are obtained through the softmax function, subsequently applied to weighted summation of the concatenated features, resulting in the final fused feature.

$$h_f = \text{softmax}(e) \times h \quad (5)$$

*Multimodal Transfer Learning*

The purpose of this step is to transfer the knowledge from the teacher model to the student model. During training, Multimodal Transfer Learning combines the teacher model's soft labels with the student model's outputs. In the experiment, we used a weighted hybrid loss function that cleverly integrates KL divergence loss with softmax cross-entropy loss. Our design enables the student model to simultaneously absorb the teacher model's prediction distribution and hard label information. The total loss (Total Loss) is calculated as shown, where α is the weighting coefficient used to balance the weights of the two losses. (Formula 6)

$$L_{total} = \alpha \left( \sum_x P(x) \log \left( \frac{P(x)}{Q(x)} \right) \right) + (1 - \alpha) \left( -\sum_i y_i \log(p_i) \right) \quad (6)$$

IV EXPERIMENT AND EVALUATION

In this section, we conduct a comprehensive evaluation of the text model and audio model (teacher model) as well as the trained fusion model (student model). Evaluation metrics include accuracy, precision, recall, and F1-score. Through multiple training rounds and parameter iterations, we optimize these metrics to enhance the optimal performance of the student model.

4.1 Evaluation Design

Our evaluation framework is designed from general to the detailed in order to address the following key research questions:

a. How does the student fusion model perform in depression classification recognition on the DAIC-WOZ dataset? (§4.4)

b. During the training of the multimodal fusion model, can the inclusion of the weights $\alpha$ and $1-\alpha$ in the combined loss function enhance the classification performance of the student model? (§4.5)

c. Does the evaluation performance of the trained multimodal student fusion model exceed that of the two unimodal teacher models? (§4.5)

d. How is the generalization capability of the multimodal student fusion model? (§4.4)

4.2 Text-Model Evaluation

In this study, we employ Logistic Regression Adaptation (LoRA) techniques in conjunction with the pre-trained Llama model within the Transformer architecture to assess its performance on text classification tasks. We define a 'DepressionDataset' class for managing the loading of textual data and associated labels. Additionally, we leverage the AutoTokenizer for preprocessing, transforming the text into a format suitable for model input, including input ids and attention masks. The preprocessing pipeline encompasses text encoding, insertion of special tokens, truncation, and padding to ensure uniformity in input length.

The text and labels are processed, batched through a DataLoader, and then passed into the model for evaluation. In the evaluation phase, the model is set to evaluation mode (non-training mode), deactivating features such as Dropout that are unnecessary during inference. The model is executed in a context that does not require gradient computation, improving efficiency and reducing memory usage.

After 20 batches of training, the model achieved an accuracy of 85.2%, a recall rate of 85.3%, and an F1 score of 85.7% (Figure 2). This indicates a good balance between accuracy and recall in the text model evaluation. To ensure the diversity of data sets and consistency of results, cross-validation is utilized during model training. Furthermore, in the section on text feature extraction, we enhanced the Llama module and added a specialized classification head to improve its capabilities in extracting text features. Originally, the Llama was primarily used for text generation and lacked advanced classification learning abilities. By integrating classification heads, we have significantly improved its ability to capture and analyze text features more accurately, enhancing the model's performance on the classification task within the DAC-WOZ dataset (see Pseudocode1). This modification not only boosts Llama's text understanding but also increases the model's accuracy in classification tasks.

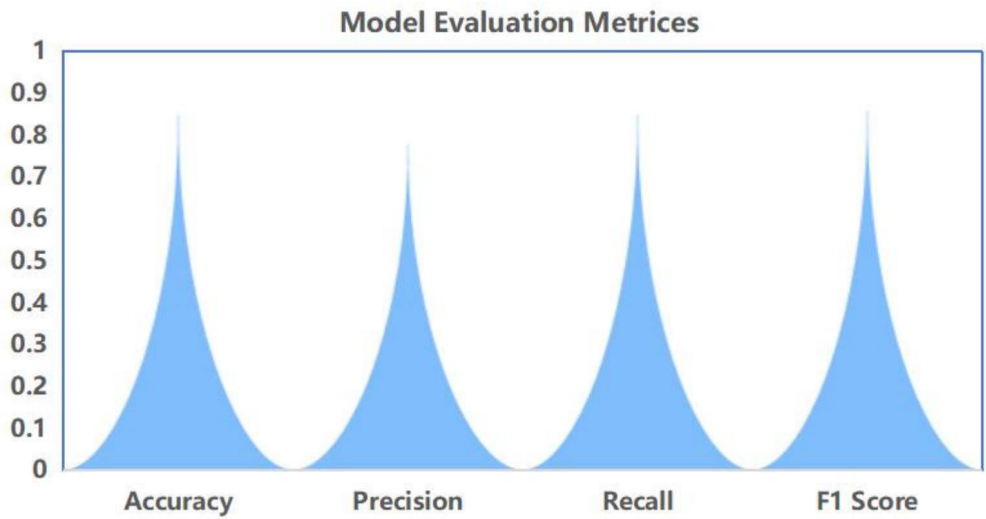

Figure2. Evaluation Metrics for Text-Model Training.

4.3 Audio-Model Evaluation

During the evaluation of the audio model, we adopted a systematic approach to verify the accuracy of the depression classification model. The evaluation was conducted on the validation set using mini-batch inference with a batch size of 4. The results yielded an accuracy of 85%, a precision of 86.6%, a recall rate of 86.5%, and an F1 score of 86.4% (see Figure 3). Experimental outcomes affirm the effectiveness of the BiLSTM algorithm in semantic understanding and classification tasks within the audio model experiments. By processing the input sequences through two independent LSTM layers (forward and backward) and concatenating their outputs, and by employing pretrained weights, the model can concurrently account for information from previous and future time steps. This enhances the ability to accurately predict mental health states from conversational audio, thereby improving modeling capability for complex sequence data.

The performance metrics of the audio model validate the robustness of our designed model against specialized datasets such as DAIC-WOZ. In actual medical scenarios, it can accurately detect mental health states from auditory cues.

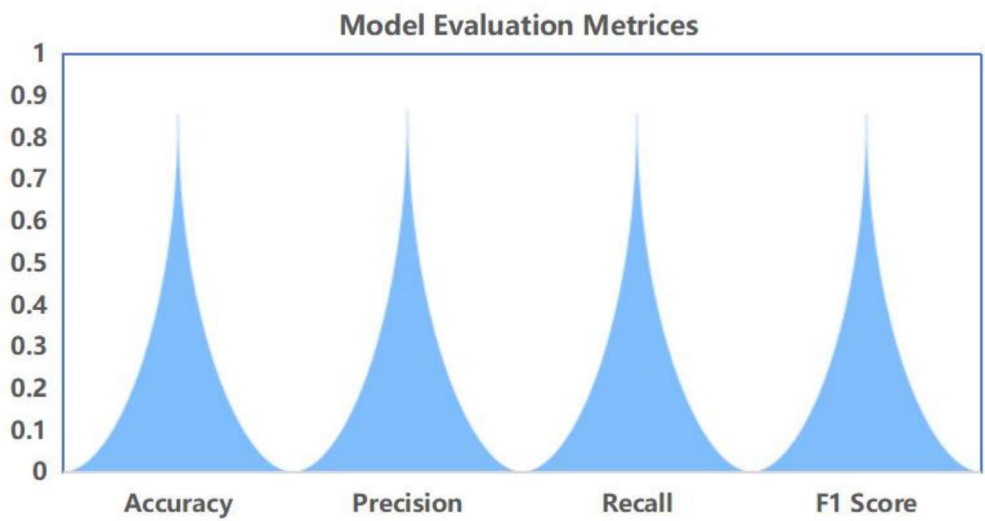

Figure3. Evaluation Metrics for Audio-Model Training.

4.4 Teacher-Student Architecture-based Model Evaluation

In the evaluation of our teacher-student architecture, we adopted a systematic approach to validate the effectiveness of the proposed student model, which integrates text and audio features from teacher models. Our work notably introduces, for the first time, a dual-teacher model fusion training methodology for a multimodal student model, which significantly enhances model performance. The evaluation process begins with data preprocessing to ensure consistency and quality of audio and text features, followed by model inference, and concludes with a comprehensive assessment using various performance metrics.

For the performance evaluation, we utilized the PyTorch framework to conduct model inference. The evaluation took place on a validation set using mini-batches of size four. We assessed the model's classification performance by calculating accuracy, weighted precision, recall, and F1 score between the model predictions and the actual labels. The experimental results indicate that the model achieved a $96\%$ accuracy rate, with precision and recall both at $96.4\%$, and an F1 score of 96.43% (see Figure 4). These metrics demonstrate not only the performance of the student model training but also a balanced sensitivity and specificity with virtually no signs of overfitting. Additionally, we calculated the Area Under the Curve (AUC) for the model, which achieved a high AUC value of $99.1\%$ (see Figure 5). This result demonstrates the model's exceptional performance in distinguishing between depressive and non-depressive states, exhibiting high sensitivity and specificity. The AUC value is a direct indicator of the model's ability to maintain a high true positive rate while keeping a low false positive rate across varying thresholds. Our model excels in handling multimodal data, particularly in merging audio and text features, effectively capturing the characteristics of data pertaining to individuals with depression.

In the context of knowledge transfer, the model transmits not the specific input data but rather the prediction results in the form of a probability distribution. Unlike hard labels, which impose sharp and definitive decision boundaries, probability distributions offer a more nuanced and continuous decision boundary. This characteristic is particularly advantageous for complex classification tasks such as depression diagnosis, where categories may overlap or exhibit ambiguous boundaries. The inherent fuzziness of probability distributions allows the model to make more robust and contextually appropriate decisions in these uncertain regions. Our experimental results on the test set, evaluated through four performance metrics, substantiate this approach. The proposed method demonstrates superior performance in the depression classification task, especially in handling multimodal data.

Based on the performance results from the test set, we particularly focused on the model's generalization ability. The experimental findings indicate that through the aforementioned four performance metrics, the superiority of our proposed method in the task of depression classification is validated. The model demonstrates excellent performance in handling multimodal data.

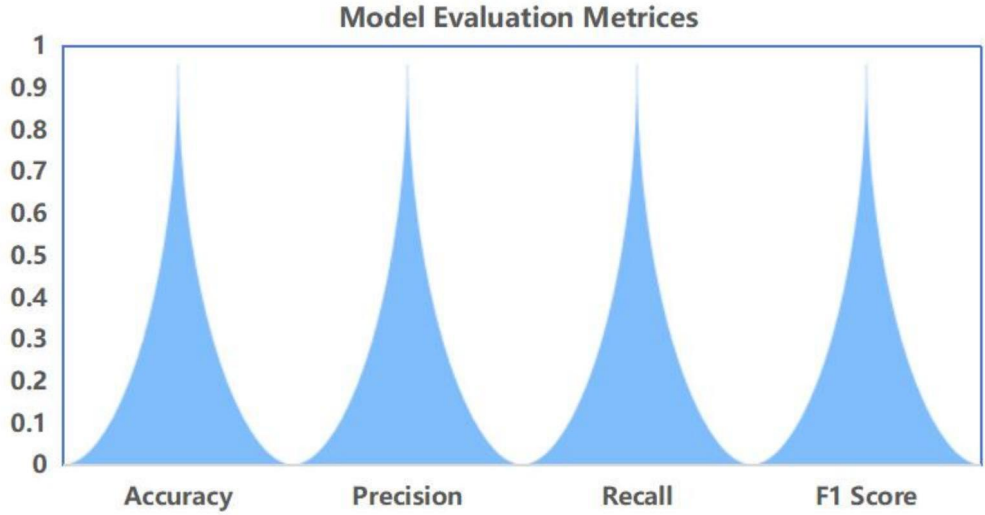

Figure4. Evaluation Metrics for Teacher-Student Architecture. This model employs two teacher models to train a multimodal student model, thereby enhancing its performance metrics. The fusion training approach results in superior overall performance of the student model.

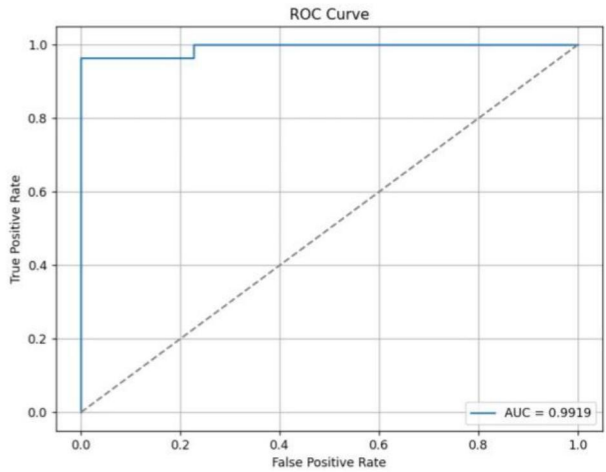

Figure5. ROC evaluation index of the Teacher-Student Architecture model.

4.5 Comparison Result

In this section, we conduct an in-depth analysis of the model through ablation studies, using the results from the DAIC-WOZ validation set as a reference. The ablation studies are divided into three main components: (1) We integrate a multi-head attention mechanism into our student fusion model based on BERT-base-uncased for feature fusion. By comparing the model performance before and after incorporating the multi-head attention mechanism, we demonstrate its effectiveness in enhancing multimodal feature integration. (2) During the training of the student model, we introduce a weighted loss function with parameters $\alpha$ and $1-\alpha$ to adjust the contributions of different tasks. This allows us to evaluate the classification performance of the final student model and validate the importance of weighted strategies in knowledge transfer. (3) We compare the final student fusion model with the original BERT-base-uncased model. Detailed ablation results are presented in §4.5.1. Additionally, we provide a performance comparison between the two teacher models (text and audio models) and the final student fusion model; these detailed experiments are outlined in §4.5.2.

*4.5.1 Ablation Experiments and Comparative analysis*

（1）Comparison of results before and after the addition of multi-head attention mechanism

In this study, we constructed a student fusion model based on the BERT-base-uncased model, and improved feature fusion through the addition of a Multi-Head Attention Mechanism. Experimental results show a significant enhancement in model performance post this addition. Specifically, the accuracy of the model increased by 4.1 percentage points, from 91.9% to 96%, the precision and recall each improved by 6.4 percentage points, rising from 90% and 94.7% to 96.4% respectively, the F1 score was elevated by 6.8 percentage points, from 92.3% to 99.1% (see Table 1).

The introduction of the multi-head attention mechanism aims to boost the model's multi-dimensional understanding of input features. Although the BERT-base-uncased model is a powerful language model, its ability to handle multimodal data (such as text and audio) through mere concatenation does not capture the complex relationships between different modalities effectively. By integrating the multi-head attention mechanism, each head focuses on features from different input modalities, thereby capturing a richer feature representation. The merged features are then fed into a classifier for prediction. This multidimensional feature extraction method allows the model to understand input data more comprehensively, thus enhancing its performance in depression classification.

Table 1. The evaluation metrics of the student fusion model based on BERT-base-uncased as the baseline model, compared to the model with the multi-head attention mechanism. The experimental results demonstrate that the multi-head attention mechanism plays a crucial role in the feature fusion process.

| Model | Accuracy | Precision | Recall | F1-Score |
|---|---|---|---|---|
| BERT-base-uncased | 86.4% | 88.9% | 84.2% | 86.5% |
| Removed the multi-head attention mechanism and weighted weights | 86.4% | 93.8% | 79.0% | 85.7% |
| Removed the multi-head attention mechanism | 91.9% | 90% | 94.7% | 92.3% |
| Student Fusion Model | 96% | 96.4% | 96.4% | 99.1% |

（2）Comparison of the model before and after adding the weighted loss function

We further delved into the potential impacts of weighted strategies on the efficacy of student models during the knowledge transfer process. The central principle involves regulating the contributions of the teacher model's soft labels and the true labels in the loss function through weight parameters.

Through meticulous adjustment of the α parameter, we successfully implemented different weight distributions between the soft labels of the teacher model and the true labels during the training phase, facilitating differentiated effects in the model learning process. The experimental data vividly demonstrate that the weighted hybrid loss function can significantly enhance the performance of the student model. Specifically, the use of a weighted loss function led to a 9.6 percentage points increase in accuracy on the validation set, reaching 96%, precision improved by

2.6 percentage points, from 93.8% to 96.4%, recall increased by 17.4 percentage points, from 79.0% to 96.4%, and the F1 score increased by 13.4 percentage points, from 85.7% to 99.1% (see Table 1). These results indicate that models with an enhanced weighted loss function significantly outperform those without in the specific task of depression classification.

In the process of knowledge transfer, the teacher model conveys knowledge through soft labels (probability distributions normalized by softmax), which include not only information about the similarity and differences between categories but also implicit feature relationships and data distribution information. This helps the student model to better understand the intrinsic structure and complex patterns of the data, rather than focusing solely on the data's details. However, if the student model relies entirely on soft labels, it may not accurately predict the true labels in some cases, as soft labels might contain noise or inaccuracies. Therefore, the design of the weight α aims to find a balance between soft labels and true labels, enabling the student model to absorb the teacher model's knowledge while maintaining accurate prediction capabilities for true labels. The experimental results further reinforce our perspective: by scientifically and rationally adjusting the α parameter, the soft labels of the teacher model can convey richer and more valuable information to the student model, thereby effectively enhancing the student's generalization ability.

（3）The comparison between the student fusion model and BERT-base-uncased model

In this set of ablation experiments, we evaluated the impact of multimodal fusion by comparing the performance of the student fusion model with the BERT-base-uncased model in depression classification tasks. The BERT-base-uncased model serves as a baseline without multimodal feature integration or knowledge transferred from teacher models. The student fusion model, which incorporates multimodal feature fusion, significantly outperforms the baseline across multiple evaluation metrics. Specifically, the accuracy of the student fusion model increased by 9.6 percentage points, from 86.4% to 96%; precision improved by 7.5 percentage points, from 88.9% to 96.4%; recall rose by 12.2 percentage points, from 84.2% to 96.4%; and the F1 score was enhanced by 12.6 percentage points, from 86.5% to 99.1% (see Table 1). These results demonstrate that multimodal fusion enables the student model to excel in depression classification tasks, significantly enhancing its classification performance.

The superior performance of the student fusion model highlights the importance of integrating multiple modalities, as it allows the model to capture complex relationships that single-modal models like BERT-base-uncased cannot detect. This capability is particularly critical in depression classification, where both textual and auditory features provide complementary insights into the condition. Therefore, multimodal fusion not only improves classification accuracy but also enhances the model's ability to generalize across diverse data sources.

*4.5.2 Comparative Experiments*

Comparison of two teacher models vs. student fusion model

We compared the performance of teacher models (audio model BiLSTM and text model Llama3-8B) with that of the student fusion model. As unimodal models, the teacher models performed well in their respective tasks. The audio model BiLSTM achieved an accuracy, precision, recall, and F1 score of 86%, 87%, 86%, and 86%, respectively, while the text model Llama3-8B achieved accuracy, precision, recall, and F1 scores of 85%, 78%, 85%, and 86%, respectively. However, in multimodal tasks, existing methods often fail to effectively integrate features from different modalities, struggling to fully utilize the complementary nature of multimodal data. In contrast, the student fusion model, incorporating a multi-head attention mechanism, is able to extract multidimensional features from multimodal data and achieve effective integration, significantly enhancing the model's classification performance. Specifically, the student fusion model reached accuracy, precision, recall, and F1 scores of 96%, 96.4%, 96.4%, and 99.1%, respectively, representing a 10%, 9.4%, 10.4%, and 13.1% improvement over the audio teacher model, and an 11%, 18.4%, 11.4%, and 13.1% improvement over the text model.

Different models process and assume data differently, leading to varied attention and feature extraction. The text model, trained on Llama3-8B, utilizes multi-head attention mechanisms to dynamically extract multi-level features, excelling at capturing long-range dependencies and global semantics. This global structure enables the model to grasp the macroscopic semantics and contextual relationships of the text. In contrast, the audio model, trained with BiLSTM, employs a bidirectional structure that emphasizes local features and extracts fixed contextual features through gradual processing, effectively capturing short-term dynamics in audio data. By integrating the capabilities of both text and audio models, the fused student model demonstrates more robust feature extraction and integration capabilities when handling multimodal data. This enables a more comprehensive understanding of the input data, achieving superior performance in depression classification tasks.

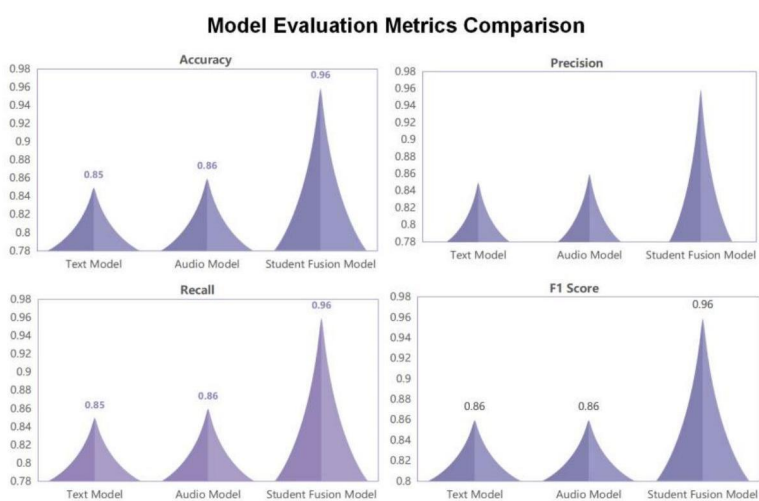

Figure 6. Performance Comparison Between Two Teacher Models (Text and Audio) and the Student Fusion Model.

Based on the multi-round comparative experiments described above, this study comprehensively validates the superior performance of the student fusion model in depression classification tasks through a systematic evaluation framework. The experimental results indicate that the student fusion model significantly enhances the model's classification performance and generalization capabilities through multimodal feature fusion and a weighted knowledge distillation strategy.

*4.6 Comparison with other approaches and models*

In recent years, traditional unimodal machine learning methods in the classification and detection of depression have faced limitations such as insufficient extraction of data features, which fails to fully utilize the complementarity of different data types, potentially leading to the omission of key features in the analysis of complex vocal information [22][23][24][25]. In contrast, our research starts from the DAIC-WOZ dataset and builds a student fusion multimodal model using deep learning methods based on features from text and audio teacher models. The model integrates textual features from the BERT-base-uncased model and audio features from the wav2vec model using a multi-head attention mechanism for feature interaction, thereby obtaining robust feature representations. Experimental results show that our designed model has good generalization capabilities and significantly enhances depression classification performance. Specifically, the student fusion model achieved optimal performance with an F1 score of 99.1%, significantly outperforming traditional methods and unimodal teacher models (audio model BiLSTM and text model Llama3-8B).

In the domain of feature extraction for models, Ming Fang et al. [22] introduced a multimodal fusion model named MFM-Att, which uses a multi-level attention mechanism to extract audio and text features and employs an attention fusion network to realize cross-modal information integration. Experimental results showed that MFM-Att outperforms other models in terms of RMSE on the DAIC-WOZ dataset. Ermal Toto's team [23] proposed a deep learning framework called Audio-Assisted BERT (AudiBERT) for the classification of depression. It enhances multimodal feature fusion through a dual self-attention mechanism by combining pretrained audio and text representation models, addressing the issue of small sample sizes in audio datasets. Our approach involves applying the attention mechanism in the feature fusion of a student fusion network model, calculating the weights of text and audio features through a multi-head attention mechanism, and obtaining a fused feature representation. This helps better understand the underlying relationships between different types of data.

Genevieve Lam et al. [24] proposed a hybrid framework that combines deep and shallow models for the assessment and classification of depression. This framework encompasses three main components: a multimodal model based on DCNN and DNN for estimating PHQ-8 scores, a model based on Paragraph Vector (PV) and SVM to infer the psychological state from interview texts, and a model based on Random Forest (RF) for classifying depression. The framework was trained and tested on the AVEC2016 dataset and results indicated a significant improvement in the accuracy of depression classification. Hanadi Solieman and others [25] utilized audio and text data from the DAIC-WOZ and DAIC datasets for model training. This included two main models: a text analysis model and a speech quality analysis model. The text model captures semantic information using pretrained word vectors (e.g., 100-dimensional embeddings) and extracts features using Convolutional Neural Networks (CNN) and Long Short-Term Memory networks (LSTM). The speech quality analysis model employs audio features (such as NAQ, H1-H2, etc.) extracted using the COVAREP toolbox, with feature extraction and classification done via CNN and LSTM networks. Both models have a sequential structure, concluding with max pooling layers and fully connected layers for final classification decisions. Compared to our model selection in the experiments, the teacher models for text and audio features were extracted using the Llama3-8B and MFCC methods and trained through the BiLSTM model. Our student fusion model used features extracted from the BERT-base-uncased model and the wav2vec model, creating a multimodal framework with BERT as the baseline during training. The final performance evaluation showed high results, demonstrating that our designed model performs better in detecting depressive samples.

Ⅴ CONCLUSION

In this study, we present a multimodal fusion model based on a teacher-student architecture, which significantly enhances the performance of depression classification tasks through the guidance of text and audio teacher models. Experimental results indicate that the final multimodal fusion student model performs exceptionally well on the DAIC-WOZ dataset, particularly when managing large-scale multimodal medical scenarios involving audio and text data classification for depression, achieving optimal performance. Through multiple rounds of comparative experiments, including unimodal and multimodal models, as well as comparisons with the basic BERT model, we validated the effectiveness of our teacher-student architecture and model fusion strategy. Specifically, our model achieved an F1 score of 99.1% on the DAIC-WOZ dataset, surpassing existing state-of-the-art methods.

Additionally, we proposed a feature fusion method based on attention mechanisms during model training. This approach maps text and audio features to the same latent space and utilizes attention mechanisms to compute feature weights from different modalities, thus achieving optimal fusion effectiveness. Experimental results demonstrate that this fusion method significantly enhances the model's generalization ability, allowing it to effectively uncover depression-related information embedded in the raw audio data.

In the future, we will further explore large multimodal language models for depression and introduce new data types, such as physiological sensor data, to enrich the model's input information. At the same time, in terms of model design, we plan to dynamically adjust the contribution of the teacher model during the training process to ensure that the student model can more effectively learn the knowledge from the teacher model. This approach is expected to further enhance the model's performance and its application scope.

## VI ACKNOWLEDGMENTS

This work was supported by the This work was supported by the Hainan Provincial Natural Science Foundation of China *Research on Landslide Recognition Technology Based on Multi-view 3D Reconstruction and Multi-dimensional Deep Neural Network from Remote Sensing Images*(grant number: 624RC529) and *University of Sanya Talent Introduction Project Oil Spill Identification on Sea Surface Using SAR Images Based on YOLOX*(grant number: USYRC22-08).

APPENDIX

---

**Algorithm 1** Data Loading Function
---
**Input: Input:** Data directory, label file, tokenizer, max length
**Output: Output:** Loaded dataset
1: Initialize empty lists: **texts**, **labels**
2: Read label file and create a label mapping
3: **for** each file in data directory **do**
4:   Extract file prefix
5:   **if** file prefix exists in label mapping **then**
6:     Read file content and merge into a single text
7:     Append text to **texts**
8:     Append corresponding label to **labels**
9:   **end if**
10: **end for**
11: Create and return a **DepressionDataset** object

---

**Algorithm 2** Evaluation Function
---
**Input: Input:** Model, test data loader, device
**Output: Output:** Evaluation metrics (accuracy, precision, recall, F1 score)
1: Set model to evaluation mode
2: Initialize empty lists: **all_preds**, **all_labels**
3: **for** each batch in test data loader **do**
4:   Move batch data to device
5:   Compute logits using the model
6:   Compute predictions (**preds**) from logits
7:   Append predictions to **all_preds**
8:   Append labels to **all_labels**
9: **end for**
10: Compute and print **accuracy**, **precision**, **recall**, and **F1 score**
11: Return evaluation metrics

---

**Algorithm 3** Main Workflow
---
**Input: Input:** Model paths, data paths
**Output: Output:** Evaluation results
1: Create tokenizer and model
2: Configure LoRA settings
3: Load data and create DataLoader
4: Call evaluation function to evaluate the model

**Pseudocode1 Text-Model Evaluation**

**Algorithm 1** Data Loading Function
---
**Input: Input:** Label file, audio file paths
**Output: Output:** Preprocessed data
1: Read the label file and create a label dictionary
2: Filter out audio files without corresponding labels
3: Filter out labels without corresponding audio files
4: **for** each audio file **do**
5:     Resample the audio
6:     Apply FIR filter
7:     Extract MFCC features
8: **end for**
9: Return the preprocessed data

**Algorithm 2** Evaluation Function
---
**Input: Input:** Model, data loader
**Output: Output:** Evaluation metrics (accuracy, precision, recall, F1 score)
1: Set the model to evaluation mode
2: Initialize lists to store predictions and labels
3: **for** each batch in the data loader **do**
4:     Get inputs and labels
5:     Perform forward propagation to obtain predictions
6:     Extract predicted classes
7:     Update the lists of predictions and labels
8: **end for**
9: Calculate accuracy, precision, recall, and F1 score

**Algorithm 3** Main Function
---
**Input: Input:** Model, data paths
**Output: Output:** Evaluation results
1: Call the **Data Loading Function** to obtain training and validation data loaders
2: Call the **Evaluation Function** with the model and validation data loader

**Pseudocode2 Audio-Model Evaluation**

**Algorithm 1** Data Loading Function

**Input:** Input: Label file, audio file paths
**Output:** Output: Preprocessed data
1: Read the label file and create a label dictionary
2: Filter out audio files without corresponding labels
3: Filter out labels without corresponding audio files
4: **for** each audio file **do**
5:     Resample the audio to target sample rate
6:     Apply FIR filter to the audio
7:     Extract MFCC features from the filtered audio
8: **end for**
9: Return the preprocessed data

**Algorithm 2** Evaluation Function

**Input:** Input: Model, preprocessed data
**Output:** Output: Evaluation metrics (accuracy, precision, recall, F1 score, AUC)
1: Set the model to evaluation mode
2: Initialize lists to store predictions and true labels
3: **for** each batch in the preprocessed data **do**
4:     Get audio inputs and transcript inputs
5:     Perform forward propagation to obtain model outputs
6:     Compute predicted classes from the outputs
7:     Update the lists of predictions and true labels
8: **end for**
9: Calculate accuracy, precision, recall, F1 score, and AUC
10: Plot the ROC curve
11: Return the evaluation metrics

**Algorithm 3** Main Function

**Input:** Input: Model path, label file, audio file paths
**Output:** Output: Evaluation results
1: Load the student model, text tokenizer, and audio processor
2: Call the **Data Loading Function** to preprocess the data
3: Call the **Evaluation Function** with the model and preprocessed data
4: Print the evaluation results

**Pseudocode3 Fusion-Model Evaluation**